\pdfoutput=1
\documentclass[10pt,twocolumn,letterpaper]{article}

\usepackage[pagenumbers]{cvpr}   
\usepackage[accsupp]{axessibility}
\definecolor{cvprblue}{rgb}{0.21,0.49,0.74}
\usepackage[breaklinks,colorlinks,allcolors=cvprblue]{hyperref}

\def\paperID{40709}
\def\confName{CVPR}
\def\confYear{2026}

\title{CrossVL: Complexity-Aware Feature Routing and Paired Curriculum for Cross-View Vision-Language Detection}

\author{
Zhipeng Liu \quad Chunbo Luo\\
Department of Computer Science, University of Exeter\\
{\tt\small \{zl481, C.Luo\}@exeter.ac.uk}\\
{\url{https://github.com/1nyourlife/Crossvl_cvpr2026}}
}

\begin{document}
\maketitle

\begin{abstract}
Vision–language models (VLMs) enable text-guided object detection but degrade severely under cross-view scenarios where ground and aerial viewpoints differ in altitude, scale, and spatial layout. These geometric changes introduce systematic complexity variations between viewpoints, e.g., ground view images contain dense and highly occluded structures, while aerial images are sparse and globally organized. Fixed VLM fusion mechanisms cannot handle this discrepancy. We propose \textbf{CrossVL}, a framework combining \textbf{Complexity-Aware Pathway Aggregation (CPA)} and \textbf{Paired Curriculum Learning (PCL)} for enhanced cross-view detection for VLM. CPA estimates scene complexity from multimodal statistics and routes visual features through multiple pathways to obtain view-specific representations. PCL leverages semantic consistency of synchronized ground–aerial pairs to provide stable early supervision and then gradually shifts toward randomized sampling. On MAVREC, CrossVL improves Florence-2’s aerial mAP from 58.66\% to 61.03\% and reduces the ground-aerial performance gap from 8.63pp to 6.65pp, while also achieving a 3.3× reduction in variance across random seeds. CPA provides stable complexity-aware feature aggregation, and PCL enhances optimization dynamics. Together, they demonstrate that coordinated architectural and training adaptations are crucial for robust cross-view VLM detection.

\end{abstract}

\section{Introduction}
\label{sec:intro}

\begin{figure}[t]
    \centering
    \includegraphics[width=\columnwidth]{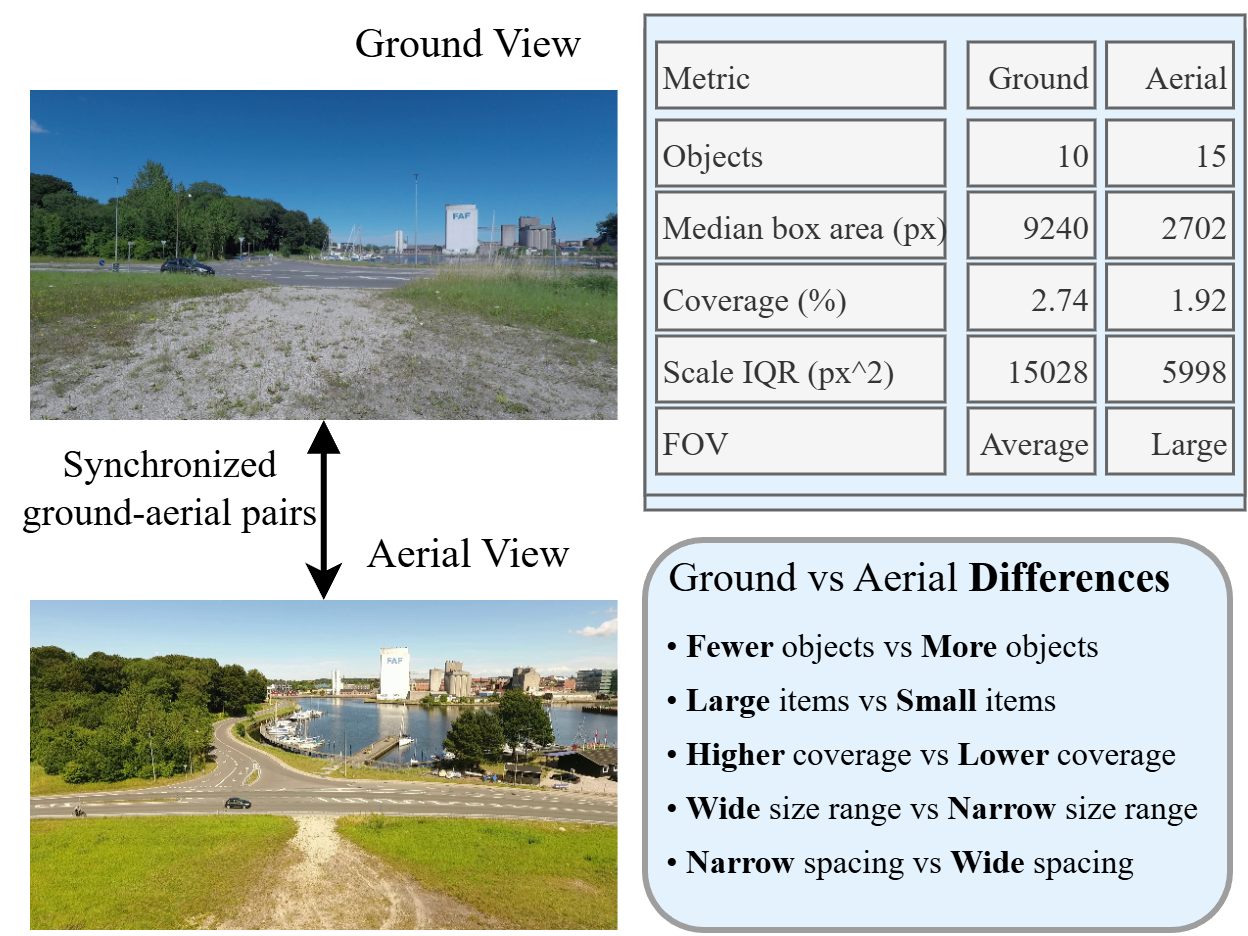}
    \caption{\textbf{Ground and aerial geometric discrepancy.}
    Sample-level statistics reveal consistent structural differences between the two views. Ground images contain fewer but larger and more closely spaced objects, while aerial images contain more objects with smaller scale, wider spatial distribution, and larger fields of view. These disparities highlight the geometry gap that challenges cross-view VLM detection.}
    \label{fig:geometry_gap}
\end{figure}

Vision-language models (VLMs) have advanced object detection by leveraging large-scale multimodal pretraining~\cite{pmlr-v139-radford21a,Li_2022_CVPR,Xiao_2024_CVPR}. Despite these gains, existing VLMs still exhibit a substantial and persistent performance degradation from ground view to aerial view under the same training protocol~\cite{Dutta_2024_CVPR}. This limits deployment in autonomous navigation, urban monitoring, and disaster response, where aerial view imagery is essential while most available training data are from ground view.

\textbf{The geometry gap.}
To understand why this gap persists even with mixed training, we examine the structural properties of synchronized ground and aerial images in Fig.~\ref{fig:geometry_gap}. Ground views contain fewer but larger objects with high coverage and tight spatial clustering, whereas aerial views contain more objects with smaller scale, broader spatial distribution, and larger fields of view. These systematic differences suggest that the core challenge arises from spatial/geometric variation rather than appearance differences.

This geometric disparity creates a fundamental \emph{complexity imbalance}: ground scenes require fine-grained processing for dense object interactions, whereas aerial scenes benefit from global contextual reasoning for sparse layouts. However, conventional VLM fusion mechanisms apply uniform processing regardless of these complexity variations, leading to suboptimal representations and training instability.

Appearance changes such as synthetic-to-real or day-to-night preserve object geometry, whereas cross-view scenarios introduce concurrent variations in viewpoint, scale, occlusion, and spatial layout. Such geometric changes make feature alignment strategies largely ineffective~\cite{Achituve_2021_WACV}. Recent analyses show that even moderate camera-parameter differences can cause notable performance degradation~\cite{Wang_2023_CVPR,NEURIPS2024_6b7e1e96}, and that geometric consistency is important for cross-view representation learning~\cite{Rhodin_2018_ECCV}. These observations reinforce that cross-view detection requires approaches specifically designed to handle geometric, not appearance, variation.

\textbf{Unexploited paired structure.}
Cross-view datasets such as MAVREC~\cite{Dutta_2024_CVPR} provide synchronized ground-aerial pairs, yet existing methods treat the two views as independent samples. Although the two views may capture the same region with non-overlapping areas, they tend to contain similar object categories and contextual cues. Such semantic consistency offers stable scene-level supervision without requiring object-level correspondence. However, this property remains underutilized in current cross-view detection pipelines.

\textbf{Our approach.}
We introduce \textbf{CrossVL}, a framework designed to address geometry-induced variation in cross-view detection. The first component, complexity-aware pathway aggregation (CPA), estimates scene complexity from multimodal statistics and routes visual features based on estimated scene complexity. The second component, paired curriculum learning (PCL), leverages the semantic consistency of synchronized ground-aerial pairs and gradually transitions from paired sampling to mixed and random sampling. This strategy provides stable early supervision and improves generalization without adding inference-time cost. Together, CPA contributes complexity-aware representations while PCL reduces optimization instability, yielding higher aerial-view accuracy and lower variance. 

\textbf{Contributions.}
This paper makes four key contributions:
\begin{itemize}[leftmargin=*, itemsep=2pt]

    \item We propose CPA, a multi-granularity fusion strategy that performs complexity-aware feature routing to accommodate ground and aerial scenes of varying complexity.

    \item We introduce PCL, a training strategy that exploits semantic consistency at the scene level in synchronized ground–aerial pairs and improves performance on aerial dataset without additional parameters or inference cost.

    \item We show that CPA and PCL provide mutual regularization benefits. CrossVL improves Florence-2-base aerial mAP from 58.66\% to 61.03\%, narrows the ground–aerial performance gap, and reduces training variance, confirming that coordinated architectural and training components outperform independent approaches.

    \item We establish a rigorous evaluation protocol for cross-view VLM fine-tuning, using strict validation-only checkpoint selection and multi-seed reporting to avoid test leakage and obtain stable and reproducible performance estimates.

\end{itemize}

\section{Related Work}
\label{sec:related}

\textbf{Cross-view and Multi-view Perception.}
Cross-view detection introduces geometric transformations that differ from conventional domain shifts. Early multi-view learning leverages geometric consistency across calibrated cameras~\cite{Rhodin_2018_ECCV, 10.1007/978-3-030-58571-6_1}, but assumes overlapping fields of view unsuitable for ground-aerial settings. Cross-view matching works focus on localization or retrieval~\cite{Hu_2018_CVPR} rather than instance-level detection.
Recent view-tuning methods address viewpoint invariance through adversarial training~\cite{Ruan_2023_ICCV} or adapter-based alignment~\cite{10.1007/978-3-031-73347-5_18}, but focus on recognition rather than detection and do not model the geometric complexity imbalance between ground and aerial views.
Our work targets robust object detection when views share temporal synchronization but have no geometric overlap.

\textbf{Vision-Language Models for Detection.}
Large-scale VLMs such as GLIP~\cite{Li_2022_CVPR}, GroundingDINO~\cite{10.1007/978-3-031-72970-6_3}, MDETR~\cite{Kamath_2021_ICCV}, and Florence-2~\cite{Xiao_2024_CVPR} have advanced open-vocabulary detection by aligning visual features with text guidance. However, these models assume consistent imaging geometry and lack mechanisms to handle structural changes in layout or object density~\cite{Wang_2023_CVPR}. Our method introduces lightweight training-time modules that improve robustness without altering VLM design or inference cost.

\textbf{Adaptive Fusion and Dynamic Routing.}
Our CPA relates to adaptive fusion in mixture-of-experts~\cite{shazeer2017outrageously}, dynamic convolution~\cite{10.1007/978-3-030-58529-7_21}, and token selection~\cite{bolya2022tome}. While these methods adapt computation based on content, they do not address geometric disparity in cross-view scenarios. CPA estimates scene complexity from multimodal statistics and routes features through pathways specifically tailored for dense versus sparse geometric regimes.

\textbf{Curriculum Learning and Paired Supervision.}
Curriculum learning~\cite{10.1145/1553374.1553380} and sampling strategies stabilize detection optimization through hard example mining~\cite{Shrivastava_2016_CVPR} and loss reweighting~\cite{Lin_2017_ICCV}. Paired supervision leverages view-synthesis consistency in stereo and video~\cite{Godard_2017_CVPR,Zhou_2017_CVPR}, but prior cross-view detection does not exploit semantic consistency in synchronized ground-aerial pairs. Our Paired Curriculum Learning leverages temporal pairing without geometric correspondence as a complementary training signal.

In summary, CrossVL contributes complexity-aware dynamic fusion for cross-view geometric disparities and paired curriculum leveraging scene-level semantic consistency. This joint design provides robustness not achieved by existing approaches through mutual regularization between architectural and training components.
\section{Method}

\label{sec:method}
\begin{figure*}[!t]
    \centering
    \includegraphics[width=\textwidth]{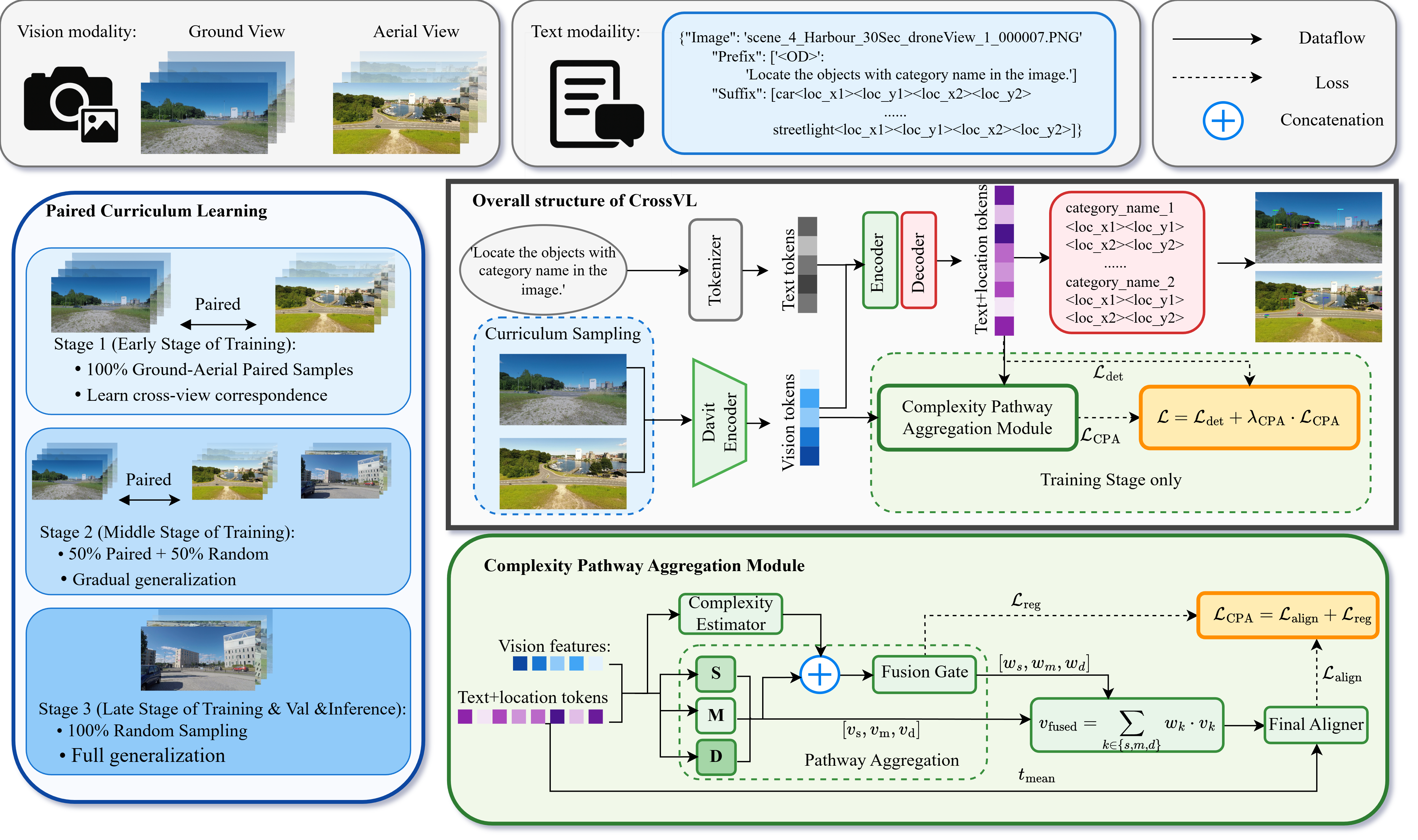}
    \caption{
        \textbf{Overview of CrossVL.}
        \textit{Left:} Paired Curriculum Learning gradually transitions training
        from fully paired to mixed and finally random sampling, leveraging semantic
        consistency in synchronized ground–aerial pairs.
        \textit{Center:} CrossVL augments Florence-2-base with CPA during training.
        CPA processes intermediate encoder/decoder features and provides auxiliary
        alignment objectives and complexity-aware representations.
        \textit{Bottom:} CPA architecture. Scene complexity is estimated from
        feature statistics and used to route visual features through sparse,
        medium, and dense pathways, followed by complexity-conditioned fusion.
        CPA is used only during training and introduces no inference-time cost.
        \textbf{S/M/D denote sparse-, medium-, and dense-pathway.}
    }
    \label{fig:framework}
\end{figure*}

\subsection{Overview}

Cross-view detection is constrained by the geometry gap between ground and aerial viewpoints: ground images typically contain fewer but larger and tightly clustered objects, whereas aerial images contain many more objects with smaller scale and wider spatial spread. This heterogeneous scene complexity destabilizes VLM fine-tuning, leading to highly view-dependent performance and large variance across seeds.

\textbf{CrossVL} addresses this challenge with two complementary components (Fig.~\ref{fig:framework}). CPA introduces complexity-aware multi-pathway processing: it estimates scene complexity from joint vision–language statistics and routes visual features through sparse, medium, and dense pathways, followed by complexity-conditioned fusion and an auxiliary alignment objective. This provides richer, complexity-aware representations during training without adding inference-time cost. PCL stabilizes optimization by structuring the training data over time: it starts from synchronized ground–aerial pairs with strong scene-level semantic consistency, then gradually transitions to mixed and finally random sampling, easing the model into harder cross-view generalization.

CPA focuses on \emph{how} features are processed under varying scene complexity, while PCL controls \emph{what} cross-view relationships the model sees over the course of training. Their interaction yields consistent gains in both cross-view accuracy and training stability.

\subsection{Problem Formulation}

We consider paired ground–aerial data
\[
\mathcal{D}=\{(I_i^{g}, I_i^{a}, y_i^{g}, y_i^{a})\}_{i=1}^{N},
\]
where $I_i^{g}$ and $I_i^{a}$ are temporally synchronized but spatially
non-overlapping views of the same location (Fig.~\ref{fig:framework}), and
$y_i^{g}, y_i^{a}$ denote their corresponding COCO-style detection labels
(bounding boxes and categories). The goal is to train a VLM capable of
\textbf{robust cross-view object detection} despite large shifts in viewpoint,
scale, occlusion, and spatial layout.

On MAVREC~\cite{Dutta_2024_CVPR}, standard fine-tuning yields strong
ground-view performance but substantially weaker aerial performance, producing a clear ground–aerial performance gap despite identical training conditions. This discrepancy reflects the underlying \emph{geometry-induced complexity imbalance} between views: dense, occluded ground scenes require fine-grained reasoning, whereas sparse aerial scenes rely more on global context. Moreover, the mismatch exacerbates optimization instability across seeds. These challenges motivate the architectural and training adaptations introduced in CrossVL.

\subsection{VLM Backbone and Data Preparation}

We adopt Florence-2-base~\cite{Xiao_2024_CVPR} as our VLM backbone,
consisting of a DaViT-3 encoder and Transformer decoder that produce structured
detection sequences given the prompt $\langle$OD$\rangle$. MAVREC's COCO-style
annotations are converted into this textual format by pairing each image with the
prompt and normalized bounding box coordinates (scaled to [0,1000]). Inference follows the standard Florence-2 decoding pipeline and COCO evaluation metrics.

\subsection{Complexity-Aware Pathway Aggregation}
\label{sec:cpa}

CPA introduces multi-granularity fusion to address the
inductive-bias mismatch between dense ground scenes and sparse aerial scenes. Ground scenes with high object density require fine-grained attention mechanisms, while sparse aerial layouts benefit from global contextual aggregation. Standard fusion architectures apply uniform processing regardless of scene characteristics, leading to suboptimal feature representations across varying geometric conditions.

\textbf{Complexity Estimation.}
CPA computes a soft complexity profile from multi-modal feature statistics:
\begin{equation}
\begin{split}
\mathbf{c} &= \mathrm{Softmax}\!\Big(
     g_\phi\big(
        [\mu(\mathbf{V}),\sigma(\mathbf{V}),\max(\mathbf{V}), \\
     &\quad\quad\mu(\mathbf{T}),\sigma(\mathbf{T})]
     \big)
   \Big) \in \mathbb{R}^{3},
\end{split}
\label{eq:complexity}
\end{equation}
where $g_\phi$ is a two-layer MLP with ReLU activation. The complexity vector $\mathbf{c}=[c_s,c_m,c_d]$ maps to sparse, medium, and dense complexity regimes. Higher feature variance $\sigma(\mathbf{V})$ typically indicates dense ground scenes with overlapping objects and complex spatial interactions, while lower variance reflects sparse aerial layouts with spatially isolated instances. The text statistics $\mu(\mathbf{T}), \sigma(\mathbf{T})$ provide complementary semantic complexity cues from the detection prompt. 

\textbf{Multi-Granularity Pathways.}
Visual features are processed by three specialized pathways (Fig.~\ref{fig:framework}, bottom) that encode complementary inductive biases tailored for different complexity regimes:

\textit{Sparse Pathway} emphasizes salient token selection through attention-based filtering. It applies a learnable attention mechanism $A_s(\mathbf{V}) = \text{Softmax}(Q_s K_s^T / \sqrt{d})$ where $Q_s, K_s$ are learned linear projections of the input features, to identify the most informative visual tokens, making it particularly effective for aerial scenes where objects are spatially isolated and globally distributed.

\textit{Medium Pathway} performs region-level aggregation using hierarchical pooling operations. It partitions the feature map into fixed-size spatial regions and applies adaptive pooling followed by cross-region communication through learnable cross-attention, capturing medium range spatial dependencies suitable for ground scenes.

\textit{Dense Pathway} captures holistic global context through full self-attention and global average pooling. This pathway excels at modeling complex interactions in densely populated ground scenes where fine-grained spatial relationships are crucial for accurate detection.

Each pathway produces a $d$-dimensional feature representation $\mathbf{V}_p$ that encodes scene information at its respective granularity level.

\textbf{Complexity-Conditioned Fusion.}
A gating module fuses pathway outputs conditioned on both pathway features and
complexity scores:
\begin{equation}
\begin{split}
\mathbf{V}_{\mathrm{fused}}
&=\sum_{p\in\{s,m,d\}} w_p\,\mathbf{V}_p, \\
\mathbf{w}
&=\mathrm{Softmax}\!\Big(
     h_\psi([\mathbf{V}_s;\mathbf{V}_m;\mathbf{V}_d;\mathbf{c}])
   \Big),
\end{split}
\label{eq:fusion}
\end{equation}
and a shared aligner projects fused visual features and text summaries into a
common embedding space.

\textbf{Training Dynamics.}
During training, CPA learns to route features based on evolving complexity patterns. Early in training, the complexity estimator produces uniform distributions across pathways, but gradually develops specialization as it learns to distinguish between sparse aerial and dense ground scenes. The routing weights $\mathbf{w}$ exhibit distinct patterns: sparse pathway dominance for aerial images and dense pathway activation for ground scenes, with the medium pathway providing smooth transitions.

\textbf{Optimization and Regularization.}
CPA is trained jointly with the VLM decoder through two lightweight objectives. 
The auxiliary vision-language alignment loss 
$\mathcal{L}_{\text{align}} = \|\mathbf{V}_{\text{fused}} - \mathbf{T}_{\text{aligned}}\|_2^2$ 
encourages consistency between fused visual features and text embeddings, providing 
a stable training signal that is less sensitive to viewpoint-induced noise. 
In addition, the routing entropy regularizer 
$\mathcal{L}_{\text{reg}} = -\sum_p w_p \log w_p$ 
promotes confident, non-uniform pathway selection and prevents collapse into a 
single pathway. These losses jointly guide CPA toward meaningful complexity-aware specialization.

Despite introducing additional pathways, CPA remains lightweight: it adds only 
2.5\% parameters relative to Florence-2-base, operates exclusively during the training 
phase, and incurs no inference-time computation or latency, making it compatible 
with real-time or resource-constrained deployment.

\subsection{Paired Curriculum Learning}
\label{sec:curriculum}

Paired Curriculum Learning stabilizes optimization by progressively increasing cross-view difficulty. Although synchronized ground-aerial pairs capture non-overlapping regions, they retain strong \emph{scene-level semantic consistency}, which provides reliable early-stage supervision. Synchronized ground-aerial pairs share environmental conditions (weather, lighting, time-of-day) and scene context even without spatial overlap, creating stable semantic anchors that facilitate cross-view feature alignment before transitioning to more challenging viewpoint generalization.

We schedule paired sampling probability as:
\begin{equation}
p_{\text{pair}}(t)=
\begin{cases}
1, & t\in[0,T_1), \\[3pt]
\text{linearly decaying}, & t\in[T_1,T_2), \\[3pt]
0, & t\in[T_2,T],
\end{cases}
\label{eq:curriculum}
\end{equation}
where $t$ denotes training progress. Early training uses paired samples to establish cross-view associations; mid training blends paired and random samples; late training uses only random sampling (i.e., uniform over all training images regardless of pairing) to promote viewpoint generalization.

\textbf{Sampling Strategy Details.} During the linear decay phase, we implement mixed sampling by randomly selecting between paired and independent samples with probability $p_{\text{pair}}(t)$. This gradual transition prevents abrupt changes in training dynamics while maintaining the semantic consistency benefits of paired supervision. The schedule parameters $T_1$ and $T_2$ are empirically set to approximately one-third and two-thirds of total training duration respectively, providing sufficient phases for cross-view association learning and generalization adaptation. We provide sensitivity analysis of $T_1$ and $T_2$ in the supplementary material, confirming robustness to schedule variations (test mAP range $<$2pp).

\textbf{Mutual Regularization with CPA.} The curriculum schedule creates a beneficial interaction with CPA's complexity-aware routing. During early paired training, semantic consistency provides CPA with stable complexity distributions, enabling reliable pathway aggregation without optimization noise. Conversely, CPA's learned complexity-aware representations prevent curriculum learning's progressive schedule from causing optimization instability, as evidenced by our stability analysis in experiments. This mutual regularization effect enables both components to achieve synergistic performance gains that exceed their individual contributions.

Together, CPA and the curriculum operate jointly during training: the curriculum shapes the cross-view sampling distribution, while CPA supplies complexity-aware representations. Their complementary design improves cross-view accuracy, reduces variance, and enables robust ground–aerial detection. Full implementation details are provided in the Supplementary Materials.
\section{Experiments}
\label{sec:experiments}

\subsection{Experimental Setup}

\begin{table*}[t]
\centering
\caption{Performance comparison on MAVREC aerial validation and test sets. Vision-only baselines from~\cite{Dutta_2024_CVPR}. *Indicates pretraining on COCO. For VLMs, we select the best checkpoint per seed based on validation mAP and report mean across 3 seeds.}
\label{tab:main_results}
\renewcommand{\arraystretch}{1.2}  
\begin{tabular}{l@{\hspace{12pt}}c@{\hspace{9pt}}c@{\hspace{9pt}}c@{\hspace{9pt}}c@{\hspace{9pt}}c@{\hspace{15pt}}c@{\hspace{9pt}}c@{\hspace{9pt}}c@{\hspace{9pt}}c@{\hspace{9pt}}c}
\toprule
& \multicolumn{5}{c}{\textbf{Validation Set (Aerial)}} & \multicolumn{5}{c}{\textbf{Test Set (Aerial)}} \\
\cmidrule(lr){2-6} \cmidrule(lr){7-11}
\textbf{Method} & \textbf{mAP} & \textbf{mAP}$_{50}$ & \textbf{mAP}$_{75}$ & \textbf{mAP}$_S$ & \textbf{mAP}$_M$ & \textbf{mAP} & \textbf{mAP}$_{50}$ & \textbf{mAP}$_{75}$ & \textbf{mAP}$_S$ & \textbf{mAP}$_M$ \\
\midrule
\multicolumn{11}{l}{\textit{Vision-only Baselines}} \\
DETR & 24.9 & 39.7 & -- & 27.6 & 45.3 & 23.6 & 40.1 & -- & 23.4 & 44.9 \\
D-DETR & 13.1 & 28.3 & -- & 14.2 & 38.1 & 10.3 & 25.0 & -- & 10.1 & 29.4 \\
D-DETR* & 31.0 & 61.7 & -- & 31.7 & 55.1 & 33.2 & 61.9 & -- & 31.5 & 51.0 \\
YOLO-NAS-L & 30.3 & 49.8 & -- & 29.2 & 61.5 & 27.0 & 43.3 & -- & 25.9 & 58.0 \\
YOLOv7 & 31.3 & 57.7 & -- & 34.2 & 61.2 & 31.9 & 58.8 & -- & 31.4 & 63.1 \\
\midrule
\multicolumn{11}{l}{\textit{Vision-Language Models}} \\
Florence-2 (random) & 63.73 & 75.22 & 68.62 & 62.08 & 49.26 & 58.66 & 71.06 & 64.19 & 59.67 & 79.81 \\
\quad + CPA (ours) & 64.49 & 76.00 & 69.58 & 61.80 & 59.66 & 60.66 & 72.61 & 66.48 & 59.51 & \textbf{85.69} \\
\quad + Curriculum (ours) & 64.37 & 76.62 & 69.63 & \textbf{63.77} & 51.92 & 56.53 & 68.40 & 61.88 & 56.20 & 81.20 \\
\quad + Both (ours) & \textbf{65.35} & \textbf{76.79} & \textbf{70.16} & 62.74 & \textbf{57.41} & \textbf{61.03} & \textbf{73.13} & \textbf{66.91} & \textbf{60.44} & 83.24 \\
\bottomrule
\end{tabular}
\end{table*}

\textbf{Dataset.}
Cross-view vision-language datasets remain limited due to the difficulty of collecting synchronized multi-view imagery with text annotations. We evaluate on MAVREC~\cite{Dutta_2024_CVPR}, currently the largest benchmark for cross-view detection. MAVREC provides 8{,}605 synchronized ground–aerial pairs for training, 538 for validation, and 1{,}614 for testing. Images span diverse urban and rural environments across northern Europe and contain 10 object categories (tram, bicycle, van, truck, bus, person, car, other, streetlight, traffic light). Each ground–aerial pair is captured simultaneously, with aerial viewpoints recorded from 25–45 m altitude.

\textbf{Baselines.}
We compare against two groups of models.
(1) \emph{Vision-only} baselines following the MAVREC evaluation protocol~\cite{Dutta_2024_CVPR}: DETR, Deformable-DETR, Deformable-DETR* (COCO-pretrained), YOLO-NAS-L, and YOLOv7.
(2) A \emph{vision-language} baseline, Florence-2-base~\cite{Xiao_2024_CVPR}, trained on randomly mixed ground and aerial images without any cross-view adaptation.

\textbf{Evaluation metrics.}
We report COCO-style~\cite{10.1007/978-3-319-10602-1_48} metrics using pycocotools: mAP (averaged over IoU thresholds 0.50–0.95 in steps of 0.05), mAP${50}$, mAP${75}$, mAP$_S$ (small objects), and mAP$_M$ (medium objects). Non-Maximum Suppression (NMS) with a 0.5 IoU threshold is applied using the Supervision library~\cite{roboflow_supervision} to remove duplicate detections. Florence-2-base does not output confidence scores; following its design, all predictions are assigned equal weight during evaluation.

\textbf{Model selection and evaluation protocol.}  
To avoid test set overfitting and ensure unbiased performance estimation, all checkpoint selection is based strictly on validation mAP. For each random seed (42, 123, 789), we train a separate model and retain the checkpoint with the highest validation performance, without accessing the test set during development. Final results report the mean and variance across these three independently trained models. Unless stated otherwise, all results are reported on the aerial validation and test sets. This strict val-only, multi-seed evaluation protocol stabilizes cross-view performance estimates and makes our comparisons more reliable.

\textbf{Implementation details.}
All experiments use Florence-2-base (230M parameters) as the backbone VLM. We train with batch size 8 and gradient accumulation of 2 steps (effective batch size 16) using AdamW with a learning rate of $1\times10^{-6}$, weight decay 0.01, and 500 warmup steps under a cosine schedule. Training uses FP16 mixed precision on a single NVIDIA RTX 5090 GPU for 10 epochs. For each seed, the checkpoint with the highest validation mAP is selected for final evaluation.

\subsection{Main Results}

We evaluate all methods following a rigorous protocol where checkpoint selection
uses only validation mAP, ensuring no test set leakage. Table~\ref{tab:main_results}
reports mean performance across 3 random seeds with their selected checkpoints.

\textbf{Vision-language models substantially outperform vision-only baselines.}
Our Florence-2 baseline achieves 63.73\% validation mAP and 58.66\% test mAP, more than doubling the performance of the strongest vision-only method  (YOLOv7: 31.3\% val, 31.9\% test). This highlights the importance of multimodal priors for cross-view detection, where geometric variations between viewpoints require semantic-level reasoning.  We also evaluated LoRA (r=16, $\alpha$=32) fine-tuning, which achieves only 24.17\% test mAP, far below full fine-tuning (58.66\%), indicating that limited parameter updates cannot capture the cross-view geometry gap.

\textbf{Individual components provide modest but complementary gains.}
CPA improves the baseline to 64.49\% validation mAP (+0.76pp) and 60.66\% test 
mAP (+2.00pp) with high stability across seeds (±1.09 std). Paired Curriculum 
improves validation mAP to 64.37\% (+0.64pp) but exhibits lower and more variable 
test performance (56.53\%, -2.13pp; ±4.97 std). CPA thus contributes stable 
complexity-aware feature adaptation, whereas curriculum learning introduces 
semantic alignment benefits but also higher optimization stochasticity.

\textbf{Combined method achieves synergistic gains in both accuracy and stability.}
Integrating both components yields 65.35\% validation mAP (+1.62pp) and 
61.03\% test mAP (+2.37pp). The validation improvement exceeds the sum of the 
individual contributions (+1.62pp vs. +0.76pp + +0.64pp = +1.40pp), illustrating 
super-additive synergy. The combined model also reduces variance by 3.3$\times$ 
compared to curriculum alone (±1.50 vs. ±4.97 std), confirming a mutual 
regularization effect: CPA prevents curriculum-induced collapse, while the 
curriculum schedule enhances CPA’s representational richness.

\textbf{Performance across object scales further highlights complementary strengths.}
CPA particularly benefits medium-sized objects (+10.4pp mAP$_M$ on validation), 
aligned with its multi-pathway feature routing, whereas curriculum learning 
improves small-object detection (+1.69pp mAP$_S$). The combined model inherits 
both strengths, delivering consistent improvements across scales and contributing 
to its overall robustness.

\subsection{Impact of Multi-Pathway Routing}

\begin{table*}[t]
\centering
\caption{Ablation study on pathway architecture. We compare the baseline, a 
single-path variant (one linear pathway without complexity-aware routing), 
and the full CPA model (three pathways with complexity-aware routing). Results 
are averaged across 3 random seeds using the best validation mAP checkpoint 
per seed.}
\label{tab:ablation_cpa}
\renewcommand{\arraystretch}{1.2}
\begin{tabular}{l@{\hspace{12pt}}c@{\hspace{9pt}}c@{\hspace{9pt}}c@{\hspace{9pt}}c@{\hspace{9pt}}c@{\hspace{15pt}}c@{\hspace{9pt}}c@{\hspace{9pt}}c@{\hspace{9pt}}c@{\hspace{9pt}}c}
\toprule
& \multicolumn{5}{c}{\textbf{Validation (Aerial)}} & \multicolumn{5}{c}{\textbf{Test (Aerial)}} \\
\cmidrule(lr){2-6} \cmidrule(lr){7-11}
\textbf{Architecture} & \textbf{mAP} & \textbf{mAP}$_{50}$ & \textbf{mAP}$_{75}$ & \textbf{mAP}$_S$ & \textbf{mAP}$_M$ &
\textbf{mAP} & \textbf{mAP}$_{50}$ & \textbf{mAP}$_{75}$ & \textbf{mAP}$_S$ & \textbf{mAP}$_M$ \\
\midrule
Baseline & 63.73 & 75.22 & 68.62 & 62.08 & 49.26 & 58.66 & 71.06 & 64.19 & 59.67 & 79.81 \\
Single-path variant (ablation) & 64.68 & 75.86 & 69.38 & 63.80 & 51.92 & 60.09 & 72.01 & 65.76 & 59.69 & 68.65 \\
Full CPA (multi-path) & 64.49 & 76.00 & 69.58 & 61.80 & \textbf{59.66} & \textbf{60.66} & \textbf{72.61} & \textbf{66.48} & 59.51 & \textbf{85.69} \\
\bottomrule
\end{tabular}
\end{table*}

To assess the effect of pathway diversity, we compare the baseline, a 
\emph{single-path variant}, and the full CPA architecture. The single-path 
variant retains only one linear visual pathway plus the final aligner, without 
using complexity-aware routing or pathway specialization.

\textbf{Single-path variant provides modest but consistent gains.}
Adding a single linear pathway improves the baseline to 64.68\% validation mAP 
(+0.95pp) and 60.09\% test mAP (+1.43pp), showing that even shallow visual-language 
alignment yields noticeable gains despite lacking multi-granularity or complexity-aware 
components.

\textbf{Full CPA further enhances generalization through pathway diversity.}
The full CPA model obtains comparable validation accuracy to the single-path 
variant (64.49\% vs.\ 64.68\%, -0.19pp) but achieves higher test performance 
(60.66\% vs.\ 60.09\%, +0.57pp). This validation-test divergence indicates that 
multi-pathway routing captures complementary receptive-field patterns that 
generalize better to aerial-view distribution shifts.

\textbf{Pathway diversity is essential for medium-sized objects.}
CPA yields its largest gain on medium-sized objects, reaching 85.69\% mAP$_M$, 
outperforming both the baseline (+5.88pp) and the single-path variant by a large 
margin (+17.04pp; 68.65\% → 85.69\%). This highlights that multi-granularity pathway 
routing is critical for handling perspective-induced scale variation. In 
contrast, small-object performance remains stable across variants 
(59.51-59.69\% mAP$_S$), suggesting that small-scale cues rely mainly on the 
shared alignment mechanism rather than pathway multiplicity.

\textbf{Routing behavior validates complexity-aware specialization.}
To verify that CPA pathways respond meaningfully to scene complexity, we analyze routing scores across scenes with varying object counts (14--179 objects per image). The dense pathway score correlates strongly with object count ($r$=0.986, $p<0.001$), while the sparse 
pathway score shows strong negative correlation ($r$=-0.988, 
$p<0.001$). This confirms that CPA successfully captures the 
complexity gradient from sparse aerial to dense ground scenes, with 
the medium pathway providing smooth interpolation between regimes.

\begin{table}[t]
\centering
\caption{Cross-view robustness on MAVREC (3-seed mean). All values in mAP(\%).
Gap = Ground - Aerial performance difference.}
\label{tab:crossview}
\small
\begin{tabular}{l@{\hspace{6pt}}ccc@{\hspace{10pt}}ccc}
\toprule
& \multicolumn{3}{c}{\textbf{Validation Set}} & \multicolumn{3}{c}{\textbf{Test Set}} \\
\cmidrule(lr){2-4} \cmidrule(lr){5-7}
\textbf{Method} & Ground & Aerial & Gap↓ & Ground & Aerial & Gap↓ \\
\midrule
Baseline     & 69.44 & 63.73 & 5.71  & 67.29 & 58.66 & 8.63 \\
+ CPA        & 70.91 & 64.49 & 6.42  & 69.96 & 60.66 & 9.30 \\
+ Curriculum & 70.46 & 64.37 & 6.09  & 69.12 & 56.53 & 12.59 \\
+ Both       & 69.08 & 65.35 & \textbf{3.73} & 67.68 & 61.03 & \textbf{6.65} \\
\bottomrule
\end{tabular}
\end{table}

\subsection{Cross-view Robustness}

Cross-view detection aims to maintain consistent performance between ground and aerial viewpoints. We evaluate cross-view generalization using 3-seed averaged results for both validation and test sets (Table~\ref{tab:crossview}).

\textbf{Baseline shows clear ground-aerial discrepancy.}
The Florence-2 baseline exhibits a notable performance gap between ground and aerial views (8.63pp on the test set). This reflects the inherent difficulty of detecting numerous small, sparsely distributed objects in aerial imagery. As expected, gaps on the test split exceed those on the validation split due to stronger viewpoint and scene-level variation.

\textbf{Individual components provide mixed effects on gap reduction.}
CPA improves both Ground (+2.67pp) and Aerial (+2.00pp) test mAP, though the larger Ground gain slightly increases the cross-view gap (9.30pp vs.\ 8.63pp). Curriculum learning increases ground-view performance but suffers from lower aerial-view consistency, resulting in the largest test set gap (12.59pp). These trends show that neither component alone fully resolves the cross-view discrepancy: CPA stabilizes complexity-induced variation, whereas curriculum learning leverages scene-level semantics yet remains prone to instability.

\textbf{Combined method achieves the strongest cross-view consistency.}
The combined approach achieves both the smallest validation set gap (3.73pp) and smallest test set gap (6.65pp), while simultaneously raising aerial accuracy from 58.66\% to 61.03\% mAP. This joint improvement—higher aerial accuracy and reduced ground-aerial discrepancy—indicates that coordinated architectural and training adaptations encourage the model to learn more view-consistent representations. The consistent reductions across both validation and test sets further support the robustness of this effect.

\subsection{Training Stability Analysis}

\begin{figure}[t]
\centering
\includegraphics[width=\columnwidth]{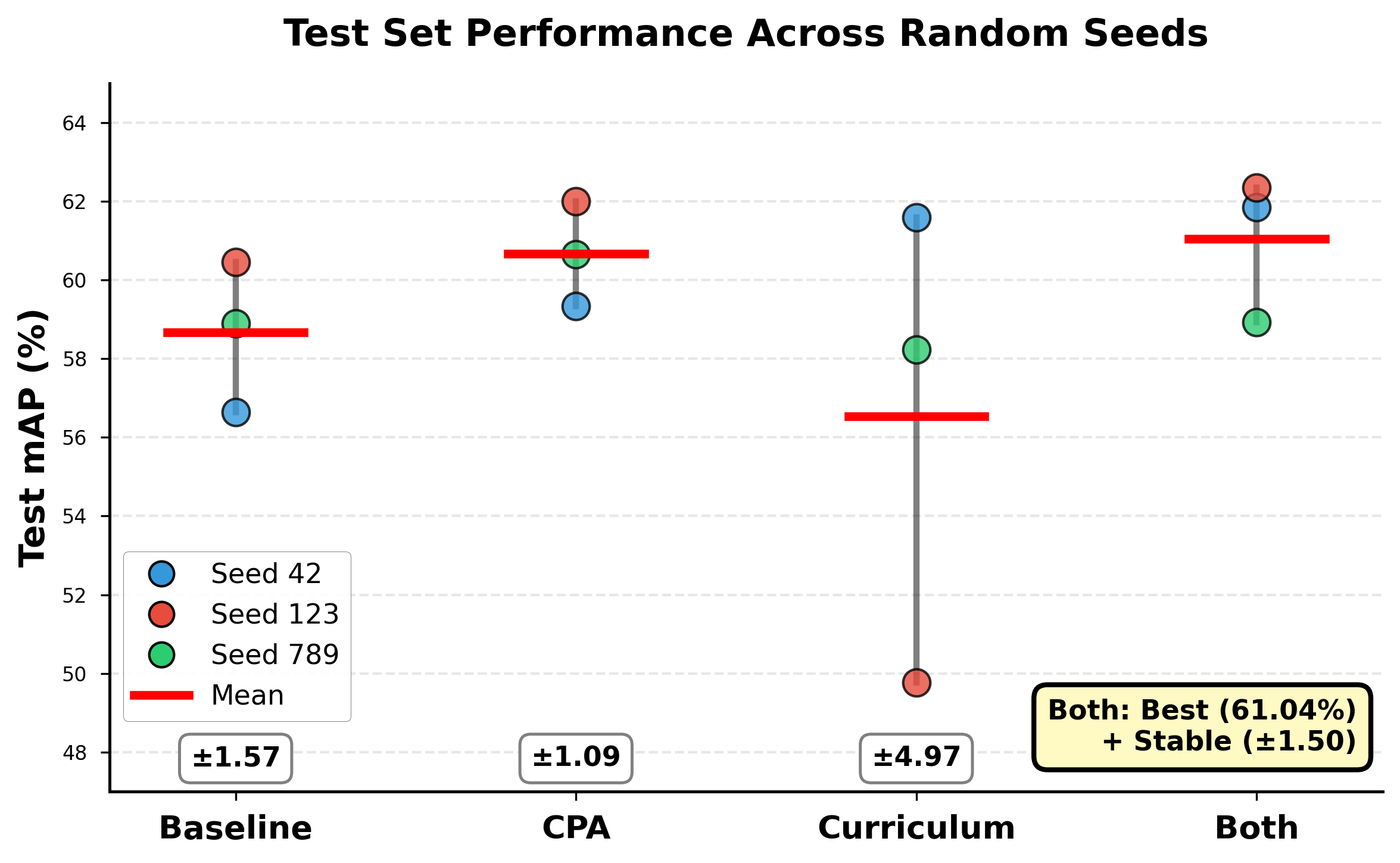}
\caption{\textbf{Training stability analysis across random seeds.} 
Per-seed test mAP distributions across three random seeds (42, 123, 789). 
Each dot represents one seed's result; red lines indicate mean performance. 
Standard deviations are shown in boxes. Our combined method (Both) achieves 
best performance with highest stability, while curriculum learning alone 
exhibits severe instability with catastrophic failure (seed 123: 49.77\% mAP).}
\label{fig:stability}
\end{figure}

Cross-view detection is highly sensitive to training instability due to the geometric gap between viewpoints. We evaluate stability across three random seeds (42, 123, 789), with Figure~\ref{fig:stability} showing the per-seed test mAP distribution.

\textbf{Curriculum learning exhibits severe training instability.}
When used independently, curriculum learning shows large variance across seeds (±4.97 std), and, critically, seed~123 encounters catastrophic failure with test mAP dropping to 49.77\%—an 11.81pp drop from the best-performing seed (61.59\%). This instability significantly limits the standalone practicality of curriculum learning despite its potential performance gains. The failure arises from the interaction between the progressive sampling schedule and random initialization, which can lead to suboptimal convergence trajectories.

\textbf{CPA provides stable training dynamics.}
CPA, in contrast, remains highly stable across all seeds (±1.09 std) while offering consistent improvements over the baseline (+2.00pp mean gain). Its complexity-aware feature routing produces robust representations that are less sensitive to initialization. 

\textbf{Combined method achieves an optimal stability–performance trade-off.}
Our integrated approach combines the strengths of both components, achieving strong performance (61.03\% mean test mAP) while reducing variance by 3.3× compared to curriculum alone (±1.50 vs. ±4.97 std). Importantly, the previously catastrophic seed~123 improves from 49.77\% under curriculum-only training to 62.34\% when CPA is included. This demonstrates a mutual regularization effect: CPA's stable representations prevent curriculum-induced failure modes, while curriculum learning enhances CPA's absolute performance.

\section{Conclusion}
\label{sec:conclusion}

We identify the geometry gap as a fundamental challenge in cross-view object detection, where concurrent changes in viewpoint, scale, and spatial layout create systematic complexity variations that existing vision-language models cannot effectively handle. Our analysis shows that this challenge differs qualitatively from standard domain adaptation, requiring approaches that explicitly target geometric rather than appearance-based discrepancies.

CrossVL addresses this challenge through coordinated architectural and training adaptations. Complexity-Aware Pathway Aggregation provides complexity-sensitive feature representations by routing visual information through specialized pathways based on estimated scene complexity, enabling effective handling of both dense ground scenes and sparse aerial layouts. Paired Curriculum Learning exploits the semantic consistency in synchronized ground–aerial pairs to stabilize early optimization before transitioning to challenging cross-view generalization.

Our most significant finding is the mutual regularization effect between these components. While individual improvements are modest, their combination yields synergistic gains and dramatically improves training stability with a 3.3$\times$ variance reduction across random seeds. The combined design also produces more consistent cross-view behavior by narrowing the ground–aerial performance gap under the same evaluation protocol. These results demonstrate that effective cross-view adaptation requires coordinated rather than independent architectural and algorithmic innovations.

\textbf{Limitations and Future Work.} The primary limitation lies in the inherent instability of curriculum learning when applied independently, as evidenced by catastrophic failures for certain random seeds (e.g., 49.77\% vs.\ 61.59\% mAP). Although the combined framework mitigates this through mutual regularization, the sensitivity to initialization suggests that more robust curriculum schedules are worth investigating. Furthermore, our complexity estimation relies on simple statistical features (mean, variance, max) that, while strongly correlated with scene complexity in our experiments (Sec.~\ref{sec:experiments}), may not generalize to scenarios where feature variance is driven by non-geometric factors such as extreme lighting or sensor noise rather than object density.

Future work should explore richer complexity indicators that incorporate spatial structure and examine the generalizability of our approach to additional cross-view scenarios beyond ground–aerial detection, such as satellite–street or indoor–outdoor viewpoints.

{
    \small
    \bibliographystyle{ieeenat_fullname}
    \bibliography{main}
}

\section*{Acknowledgements}
This project was partially supported by Horizon Europe Grant No. 101236387. Zhipeng Liu was sponsored by the joint China Scholarship Council (CSC)–University of Exeter studentship.

\end{document}